\documentclass{article}
\usepackage[dvipsnames]{xcolor}
\usepackage{sigbovik2021_conference,times}
\PassOptionsToPackage{hyphens}{url}
\usepackage[hidelinks]{hyperref}
\usepackage{url}
\usepackage{CJKutf8}
\usepackage{soul}
\usepackage{dirtytalk}
\usepackage{epigraph}
\usepackage{subfig}
\usepackage{pifont}
\usepackage{amssymb}
\usepackage{listings}
\usepackage{amsmath}
\usepackage{graphicx}

\usepackage{pgfplots}
\pgfplotsset{width=15cm, height=7cm,compat=1.8}

\definecolor{codegreen}{rgb}{0,0.6,0}
\definecolor{codegray}{rgb}{0.5,0.5,0.5}
\definecolor{codepurple}{rgb}{0.58,0,0.82}
\definecolor{backcolour}{rgb}{0.95,0.95,0.92}

\lstdefinestyle{pystyle}{
    backgroundcolor=\color{backcolour},   
    commentstyle=\color{codegray},
    keywordstyle=\color{codegreen},
    numberstyle=\tiny\color{codegray},
    stringstyle=\color{codepurple},
    basicstyle=\ttfamily\footnotesize,
    breakatwhitespace=false,         
    breaklines=true,                 
    captionpos=b,                    
    keepspaces=true,                 
    numbers=left,                    
    numbersep=5pt,                  
    showspaces=false,                
    showstringspaces=false,
    showtabs=false,                  
    tabsize=2
}

\title{
On the Origin of Species of \\
Self-Supervised Learning
}

\author{Samuel Albanie,
Erika Lu, 
Jo\~{a}o F. Henriques \\
Artificial Naturalist Society\\
Our childhood bedrooms\\
The Amazon (no, the other one)\\
} 

\iclrfinalcopy 

\begin{document}
\maketitle

\begin{abstract}

% \footnotetext{\hspace{-0.15cm}*Lexicographic genome order was used to sort the authors (first base-pairs A-T, G-C and T-A, respectively).} 

In the quiet backwaters of cs.CV, cs.LG and stat.ML, a cornucopia of new learning systems is emerging from a primordial soup of mathematics---learning systems with no need for external supervision.
To date, little thought has been given to how these \textit{self-supervised} learners have sprung into being or the principles that govern their continuing diversification.
After a period of deliberate study and dispassionate judgement during which each author set their Zoom virtual background to a separate Gal\'apagos island,
we now entertain no doubt that each of these learning machines are lineal descendants of some older and generally extinct species.

We make five
contributions:
(1)~We gather and catalogue row-major arrays of machine learning specimens, each exhibiting heritable discriminative features;
(2)~We document a mutation mechanism by which almost imperceptible 
changes are introduced to the genotype of new systems, but their phenotype (birdsongs in the form of tweets and vestigial plumage such as press releases) communicates dramatic changes;
(3)~We propose a unifying theory of self-supervised machine evolution and compare to other unifying theories on standard unifying theory benchmarks, where we establish a new (and unifying) state of the art;
(4)~We discuss the importance of digital bio-diversity,
in light of the endearingly optimistic Paris Agreement.\footnote{Our remaining contribution was charitable rather than scientific, for tax reasons.}

\end{abstract}

\epigraph{All models are wrong, but some will win you a Kaggle competition.}{George E. P. Box,\\\emph{Science and Statistics, 1976}}

\section{Introduction} \label{sec:intro}

The \textit{Great Bidecade of Annotation}\footnote{
A term muttered by bards, poets and makars in hushed tones
to describe the era 2000-2020 AD
as they queue patiently,
separated by appropriate intrinsic British emotional and social distancing measures,
for the re-opening of Will's Deli.
} has supplied humanity with vast quantities of labelled sensory data. Uncomfortably large strides forward have been taken in foundational computer vision tasks, yielding algorithms that can segment biological cells, objects, actions and IKEA folding chairs against the challenging backdrop of a minimalist Scandinavian kitchen~\citep{flyingchairs}.
A key challenge in scaling these successes to other important tasks---ultimately including non-Euclidean signals in non-Scandinavian kitchens---is that obtaining such annotation is extremely costly (and hard to assemble).

\begin{figure*}[t]
  \centering
  \includegraphics[clip,trim={0cm 5cm 2cm 3.5cm},width=\textwidth]{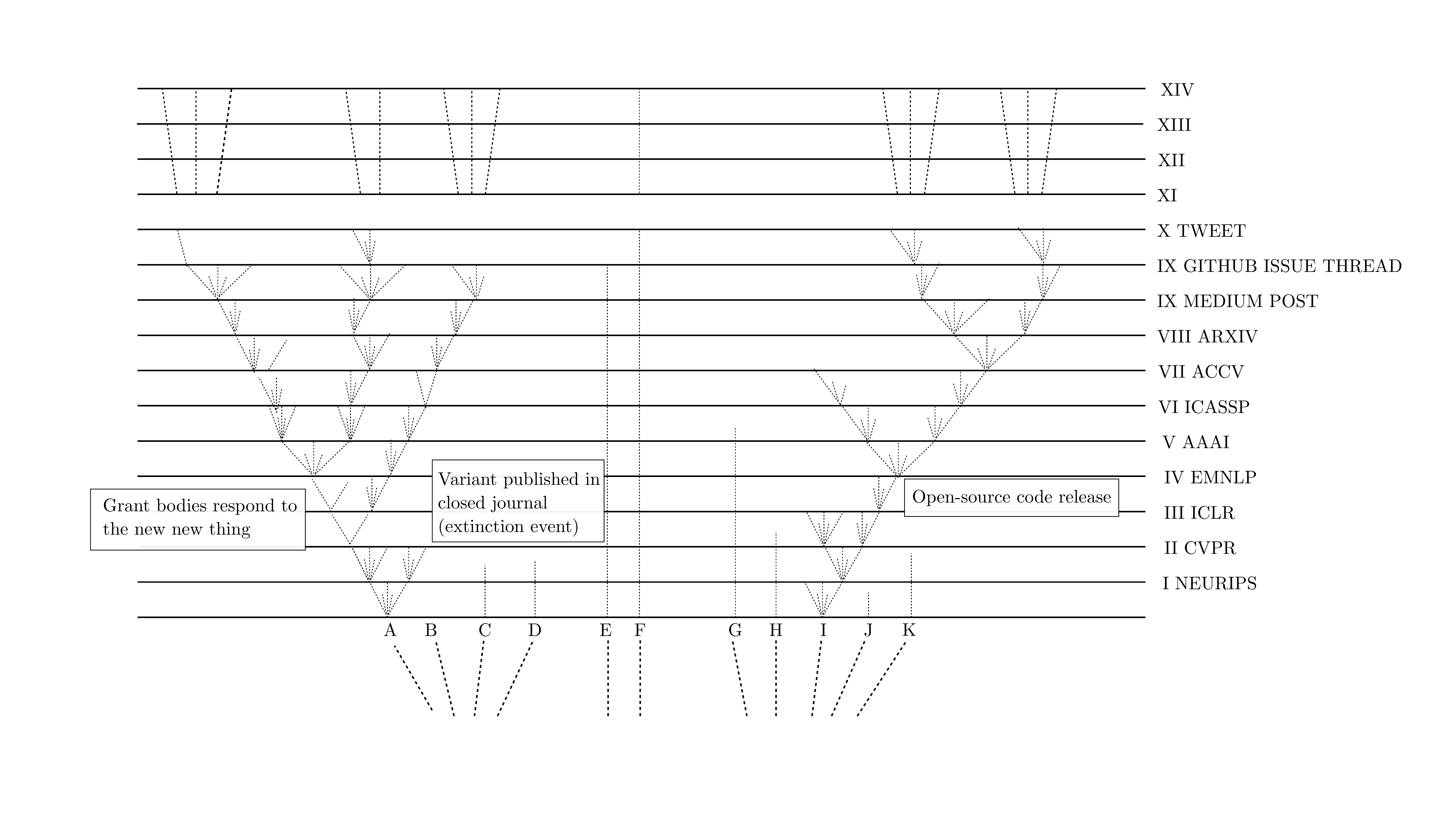}
  \caption{\textbf{Development of self-supervised learning}. Letters A through K denote self-supervised learning species in a machine learning genus, whose evolution is depicted across many generations. The intervals between horizontal lines denote the formation of large numbers of algorithmic variants over time. Horizontal lines themselves reflect examples of generational markers at which distinguishing traits can be identified using the sentence that begins ``Unlike prior research...'' in the related work sections of corresponding papers. They also serve to improve the gestalt of the figure. We note a remarkable resemblance to the diagram presented in~\cite{darwin1859origin}.
  Letter G shows the fate of DODO, an early expert system.
  Letter F shows an as yet unpromising research direction stubbornly pursued by an isolated professor over the ages, sometimes referred to as a \emph{living fossil}.
  \label{fig:tree-of-life}
  }
\end{figure*}

One promising solution lies in a niche but growing breed of machine autodidactism known as \textit{Self-Supervised Learning} (SSL).
With the potential for reduced teaching expenses and a secure acronym, this approach engages the machine in a profitable ``self-education'' exercise to render it maximally useful for a given downstream career path.\footnote{We note that today's neural networks, after training and being deployed to a professional environment, do not sufficiently engage in on-the-job learning, and thus have their career growth significantly curtailed. This will be discussed in an upcoming article in the journal \emph{American Sociological Review}, pending the successful crossing of the Atlantic Ocean of our manuscript by steamer.}
However, despite its clear cost-cutting benefits and notable impact to date, 
little is known of the origins of this behaviour in the machine education establishment. 

As classically trained machine naturalists aboard HMS Arxiv, we were much struck with certain facts in the distribution of self-supervised learning machines, and with the relationships of the loss functions of the present to those of the past.
These facts seemed to us to throw some light on the origin of species of self-supervised machines---that ``mystery of mysteries'', as it is already referred to by our greatest
stoic
Twitter philosophers.\footnote{When told (@mentioned) about our discoveries, Seneca replied: ``Cool.'' Brevity is the soul of wit.}

In this work, we report our findings, structuring them as follows. After strengthening our novelty with references to questionably applicable literature (Sec.~\ref{sec:related}), and ignoring one reference in particular, we
then sensitively explore that most savage of topics, the \textit{Struggle for Existence}, and examine its role within a framework of \textit{Unnatural Selection} of the fittest self-supervised learning machines (Sec.~\ref{sec:framework}).
We then evaluate the resulting unifying theory on competitive unifying theory benchmarks, where we demonstrate a generational advance over prior state of the art (Sec.~\ref{sec:experiments}). We conclude abruptly (Sec.~\ref{sec:conclusion}).
\section{Related Work \label{sec:related}}

Our work builds architecturally unsound bridges between two appropriately disconnected themes in the literature:
(i) \textit{the development of self-supervised learning} and
(ii) \textit{grand unifying theories}.

\textbf{The development of self-supervised learning.} The benefits of self-supervised pedagogy have been known to \textit{homo sapiens} since the scholarly efforts of~\cite{tufail}, who showed that are there are few limits to what a self-directed intellect can achieve when it brings to bear the kind of calm, phlegmatic reasoning that determines that dissecting your recently deceased adopted mother will be an instructive exercise. %\footnote{True story.}
A string of autodidact successes followed, with 
the steamy patents of socialite James \say{turn down for} Watt,
the number theory wizardry of conscientious Ramanujan\footnote{\say{I like big integers and I cannot lie.}},
the soul-moistening licks of Django Reinhardt
and the insta-translations of Kató \say{babel fish} Lomb.
Despite its auspicious and well-publicised start among humans, however, little is known of the origins of this behaviour in the machine education establishment. To address this, we initiated a search, starting in international territory
and lawless dark-web waters,
with a careful examination of specimens across publicly accessible global pre-print servers. As the search grew, we encountered the walled kingdoms of JSTOR and ScienceDirect and carefully obtained VPN visas to ensure safe passage deeper into the academic wilderness.

Surveying the landscape, we first encountered specimens of related, but quite distinct species of \textit{self-organising} maps~\citep{von1973self,kohonen1982self}, \textit{self-interested} agents~\citep{barto1985learning} and \textit{self-learning} controllers~\citep{nguyen1990neural}. After discovering a general self-supervised framework for reinforcement learning that established a new benchmark for creative figure artwork~\citep{schmidhuber1990making}, we came upon the work of~\cite{de1994learning} that popularised the use of self-supervised representation learning through cross-modal hypothetical bovine prediction.
Hacking further into the unkempt forest, barely visited by journal surveyors, our earliest finding was a self-supervised algorithm for the task of Telugu vowel recognition, creatively coupling adaptive learning with fuzzy set membership. Upon encountering new samples, this algorithm would assign estimated class memberships to those that fall close to existing sample clusters and iteratively re-estimate model parameters with the updated assignments~\citep{pal1978computer},
which is clearly too much work when falling back to preconceived notions will do just as well.

Exhausted from clicking on Google Scholar listings that failed to link an accessible PDF, we paused to rest and taken on water. We had about 80 open browser tabs consuming a total of 48GB of RAM, and a handful of clues hinting at parallel, independent algorithmic isolated germinations rather than a monogenistic narrative. With our greatly diminished purses, we lacked the funds to conduct an effective \textit{alltagsgeschichte} study to establish further facts, and we thus turned to that bastion of science, the grand unifying theory, to weave together our threads into a rigorous origin story.

\textbf{Grand unifying theories.}  The history of science is strewn with courageous efforts from big-picture thinkers, unhappy with the limiting confines of the existing picture frame.\footnote{A notable example was the move from 4:3 to 16:9 aspect ratio.}
After earlier stargazers had laid the groundwork~\citep{nubians4800}, 
Babylonian astronomers were first to publish (in peer-reviewed cuneiform on sufficiently durable clay) a unifying theory tackling the periodic behaviour for the celestial bodies~\citep{babylonian1700} in their time off from innovative horticultural construction projects. The philosophical foundations of numerical analysis were then established by~\cite{iching} with \begin{CJK*}{UTF8}{bkai}易經\end{CJK*}, and household Greek names soon followed with grand theories of atoms~\citep{democritus} and axioms~\citep{archimedes}, works which remain influential even today~\citep{aaronson2013quantum}.
Apple enthusiast, amateur bodkin opthalmologist and all-round scientist extraordinaire~\cite{newton} laid massive foundations for modern science many years later with a theory that neatly pulled together the prior efforts of Kepler, Galileo and Granny Smith. Following further unifying improbable insights~\citep{laplace1829essai} and attractive analysis~\citep{maxwell1865}, the establishment batting average consequently looked commendable approaching the latter half of the 19th century. Indeed, with the acute success of the Erlangen program to unify geometry~\citep{klein1872vergleichende} and an organic treatise on natural selection~\citep{darwin1859origin},\footnote{Note: Reviewer one suggested that Darwin restrict his focus to pigeons, ``which are of interest to everybody.'' We've all had that.} the rose-tinted lens of history has prepared for us a unifying narrative in need of no further Instagram filter.  

Mother nature, though, was far from ready to lay her hand on the table, and the cracks soon began to appear in the middle order. Despite diagrams that work well for T-shirt designs, the grand hypothesis \textit{Ontogeny recapitulates Phylogeny} of~\cite{haeckel1866generelle} needed more development.
Next, logicians' logician~\cite{hilbert1922} commuted in with a plucky but ultimately unsuccessful program to prove the consistency of mathematics.
Little need be said about the
respective innings on quantum gravity
of 
those most dependable of opening batsmen,
Einstein and Schr\"odinger.
And then of course, there is string theory, the notably \rule{1cm}{0.15mm}\footnote{Insert positive/negative term according to personal preference. Since even the Wikipedia page isn't quite sure, we follow the fiscally prudent approach espoused by~\cite{shtetl06}. It is also a neat coincidence that the placeholder looks like a string.} mathematical formulation of everything.
Science, it seems, may be on the back foot, peering upwards with worried visage towards the unified heavens\footnote{Sensibly checking for cloud cover, since any application of Duckworth–Lewis-Stern at this stage of play spells crushing defeat.}.
Standing at the crease of an increasingly strained cricket analogy, it faces a doosra: is the modern scientific endeavour doomed to stand forever trembling in the shadows of those $20^{\mathrm{th}}$ century titans who swung gloriously for the boundary but came up short?\footnote{We note that recently, several adventurers have declared new efforts at unifying theories of physics~\citep{wolfram20,weinstein2020}. It seems difficult. We wish them well.} %\samuel{Theory of Gyres}

Yet the chance remains that the program of our universe may still prove itself to be a \textit{short one}, dovetailed amidst a myriad of longer alternatives by the Great Programmer~\citep{schmidhuber1997computer}. And so, 
tired, hungry, convinced that its next squeeze from the toothpaste tube really might be the last, the quest for grand theories nevertheless lives on.
Thus, undeterred,
we add our diminutive shoulders to the wheel,
recant unconventional anger management advice~\citep{thomas51},
and, in Sec.~\ref{sec:framework},
lay down our plans for a new, grand and unifying theory.

Following best-practices established in the alphabetically-related work proposed by~\cite{fouhey2012kardashian}, we conclude our literature review by citing highly original work that is related to ours by title prefix string, viz.
\textit{On the Origin of Money}~\citep{menger1892origin},
\textit{On the Origin of Speech}~\citep{hockett1960origin},
\textit{On the Origin of Objects}~\citep{smith1996origin},
\textit{On the Origin of Orogens}~\citep{jamieson2013origin},
\textit{On the Origin of Heterotrophy}~\citep{schonheit2016origin},
\textit{On the Origin of Neurostatus}~\citep{kappos2015origin} and
\textit{On the Origin of Species by Means of Natural Selection, or the Preservation of Favoured Races in the Struggle for Life}~\citep{darwin1859origin}.
Widely considered to be a cult classic, Darwin's \textit{Origin of Species} franchise is set to be rebooted for modern audiences with the gritty prequel film \textit{Gal\'apagos Origins: Warbler Finches 1835}.
\begin{figure}
    {\centering
    \includegraphics[height=4cm]{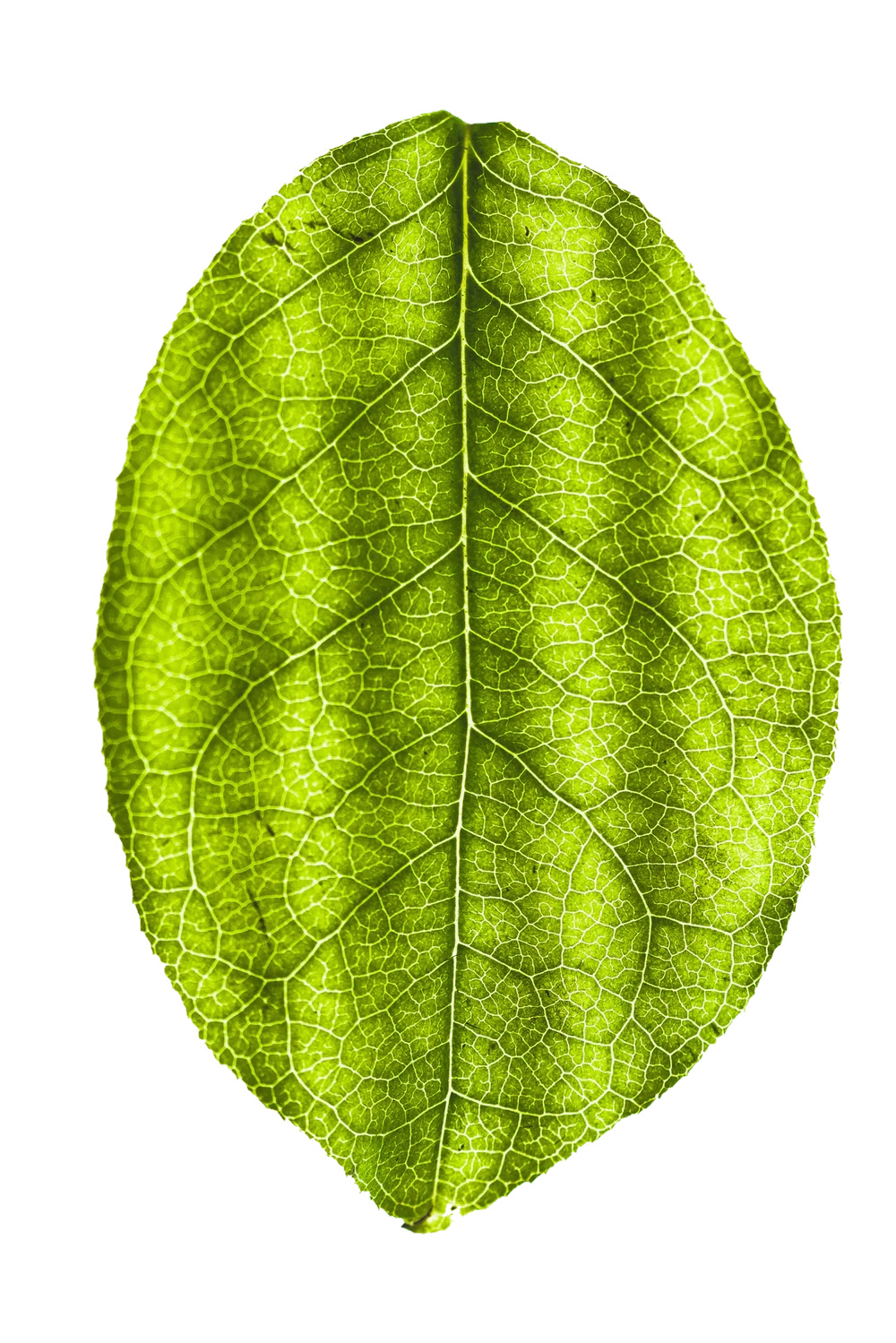}\hspace{1em}
    \includegraphics[height=4cm]{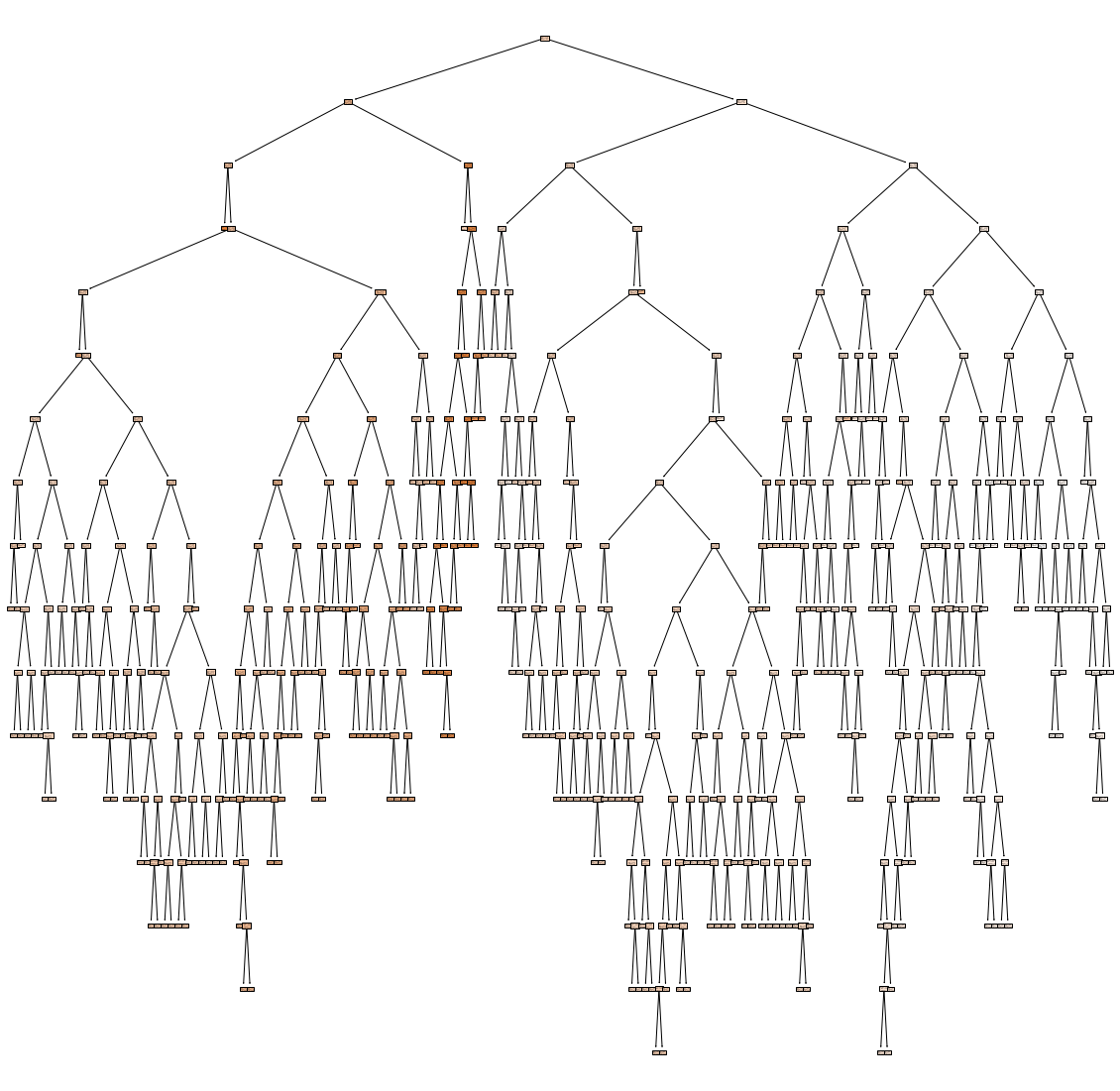}\hspace{1em}
    \includegraphics[height=4cm]{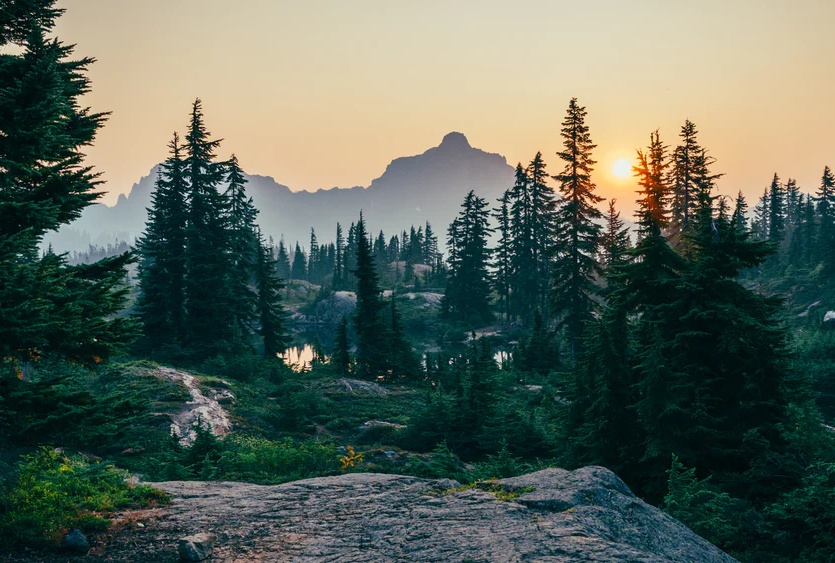}\hspace{1em}}\\
    \hspace*{1cm}(a)\hspace{3.5cm}(b)\hspace{5cm}(c)
    \caption[Caption for LOF]{A multi-level analysis of several machine learning data structures found \emph{in the wild.} (a) A random leaf. (b) A random tree. (c) A random forest. It is important not to miss (c) for (b).\footnotemark{}}%
    \label{fig:forest}
\end{figure}

\footnotetext{Image credits:~\citet{leaf,tree,forest}.}

\setlength{\epigraphwidth}{6.5cm}
\epigraph{Find someone who looks at you the way the way VGG-16 looks at a $3\times224\times224$ uint8 pixel array.}{Old English Proverb}

\section{Unifying Theory} \label{sec:framework}

Given a set of sensory measurements in some appropriately spacious collection, $\mathbf{x} \in \mathcal{X}$, self-supervised learning proceeds through a mathematical game of \textit{hide-and-seek}. First, a hiding function $h: \mathcal{X} \rightarrow \mathcal{X}$ identifies some characteristic or attribute of $x$ and hides it under a hat, $h(x) = \hat{x}$.
The $x$ is still visible in our notation for illustration purposes only.
It then falls to the seek function, $s: \mathcal{X} \rightarrow \mathcal{X}$ to find what was hidden and recover $x$,
$s(\hat{x}) \approx x$.
Since it's just a game after all, $s(\cdot)$ agrees to lose by $l(s \circ h \circ x, x) \in \mathbb{R}$ to the degree that she fails to reconstruct $x$ accurately.
So far, so simple. And yet, at the time of writing, millions of such games are being played on increasingly warm silicon across the globe, each with its own subtle tweak to the rules of the game, the stature of the players and the measurements with which they play. How did we get here? Paraphrasing Enrico Fermi, ``Where are they (the creators of these marvellous creatures)?''

To address this question, we first conducted a study of the \textit{variation in self-supervised learning}.
Inspired by the findings,
we then propose a \textit{unifying theory for the origin of self-supervised learning}.

\textbf{Variation in Self-Supervised Learning.} We began our study of variation within the academic lab, where we observed significant differences in learning system architectures emerge through the idealistic and hopeful designs of first year PhD students. We passed next to the variation found in the open landscape of the \textit{academic wilderness}, populated by papers from an exotic jungle of sources:
the wild-eyed late-stage graduate student in the throes of a final thesis push,
the wizened postdoc (now studying their fourth language),
the industrial research scientist (whose relaxed smile exudes confidence in their health insurance),
the independent researcher (too maverick to fit inside the system, too creative to throw in the research towel), 
the startup warrior (battling the manuscript as the runway crumbles beneath them) and the tenure-track professor (just 2.3 years away from her next night of sleep). Here too, we found an abundance of variety at every turn (see Fig.~\ref{fig:forest} for examples).  
Digging deeper, we studied fossil evidence
from a number of 90's webpages in University servers which have been isolated for decades, lacking any inbound hyperlinks from the wider internet.
It was here that we made a striking discovery:
the dramatic phenotype changes in chirps and vocalisation patterns in the tweetverse, and vivid colours of visualisations in blog posts, were all the result of imperceptible source code (genotype) changes induced by a novel mutation mechanism.

\noindent \textbf{Unnatural Selection:
A Unifying Theory for the Origin of Self-Supervised Learning.}
Excited by our discovery, we sought to better understand this mutation effect and observed the following:
It is widely known that
the primary mechanism by which a new codebase is formed is by combining the top two methods on paperswithcode.com to eke out a 0.001\% mAP improvement. Crucially, however, reproduction of results from an identical \texttt{git clone} is not guaranteed, due to external conda environment factors such as rainforest humidity levels.

Since the resulting diversity is produced in a competitive \textit{publish or perish} environment, a struggle for existence then ensues, pruning species that do not wish to be pruned.
Over generations, the variety produced by this process, termed \textit{unnatural selection}, can be tremendous (we visualise this effect in Fig.~\ref{fig:tree-of-life}).

The implications of this theory are profound. For many centuries, scholars have been perplexed by the complexity of ``research code'' found in the wild. Through unlikely combinations of Stack Overflow snippets, strangely fortuitous bugs and haphazard merges of git conflicts, these projects would produce publishable results despite defying all known laws of Software Engineering. The traditional dogma put this down to the designs of an all-knowing Supervisor.
Yet the evidence we have gathered now suggests it to be instead a process of gradual diverging changes from previous codebases, back to a hypothesised ``Initial commit'' in an SVN repository eons ago.
We can only speculate about unknowable protozoal generations of e-mailed zipped snapshots of even earlier versions.

\section{Experiments} \label{sec:experiments}

In this section, we comprehensively validate our theory with \textit{in carbono} experiments.
Given the rapid rate of reproduction of self-supervised learning systems, we were able to follow the example of monastic-fantastic Gregor \cite{mendel1865versuche} and his famous pea-breeding experiments (as part of our 5-a-day), and enlarged the scope of our experiments to geological timescales, encompassing 1,000 generations of proposed systems, or about one week of arXiv submissions.

From a modest initial population composed of nothing but support vector machines and fuzzy logic models (a protected species at risk of poaching due to its luxurious fur), we observed a cornucopia of methods emerge: gradient descentipedes large enough to fit a standard full-page figure; colonies of cross-antropy loss functions;
angiosperm plants with copious pollen-omial production;
mothogonal initialisers;
cicadagrad (with very noisy gradients);
and beartypes (which are much stricter than their equatorial python counterparts~\citep{beartype}).
These specimens were capable of multiple tasks of the natural world, such as se-mantis segmentation or trans-fur learning.
As our awareness of the growing absurdity of the number of puns also grew, we decided to hide in a nearby Random Forest and narrate from a Conditional Branch with a very on-the-nose impression of Sir David Attenborough.
It was obvious that we were sitting precariously close to the front of a feedforward food chain, and we did not want to personally test whether we still enjoyed humanity's status as apex predators, or had been downgraded to prey.
We decided to shield ourselves on the Rainy Picnic Corollary of the No Free Lunch Theorem, and returned home in time for (pea-based) supper.

\begin{figure}
    \centering
    \includegraphics[height=0.31\textwidth]{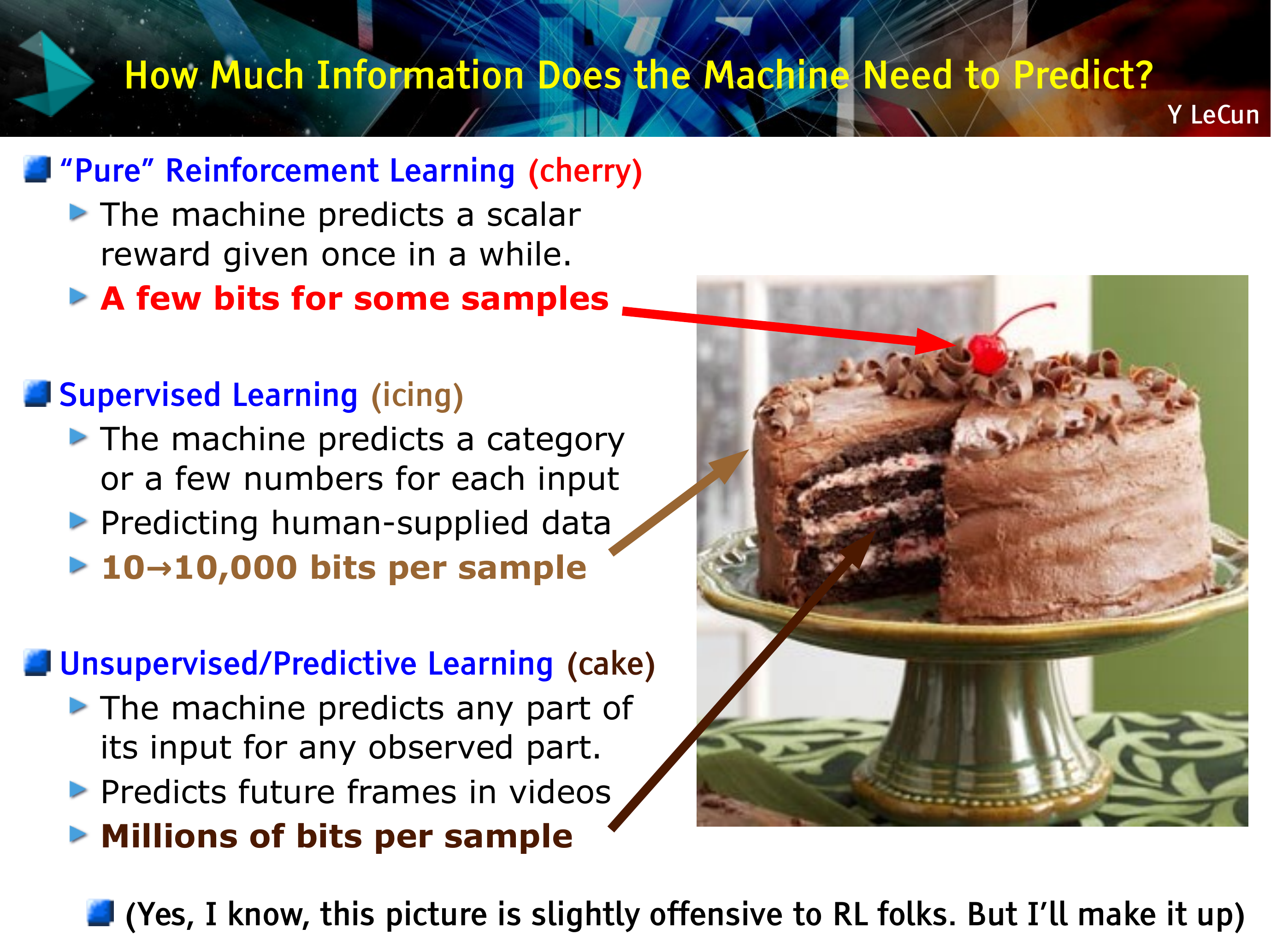}%
    \includegraphics[height=0.31\textwidth]{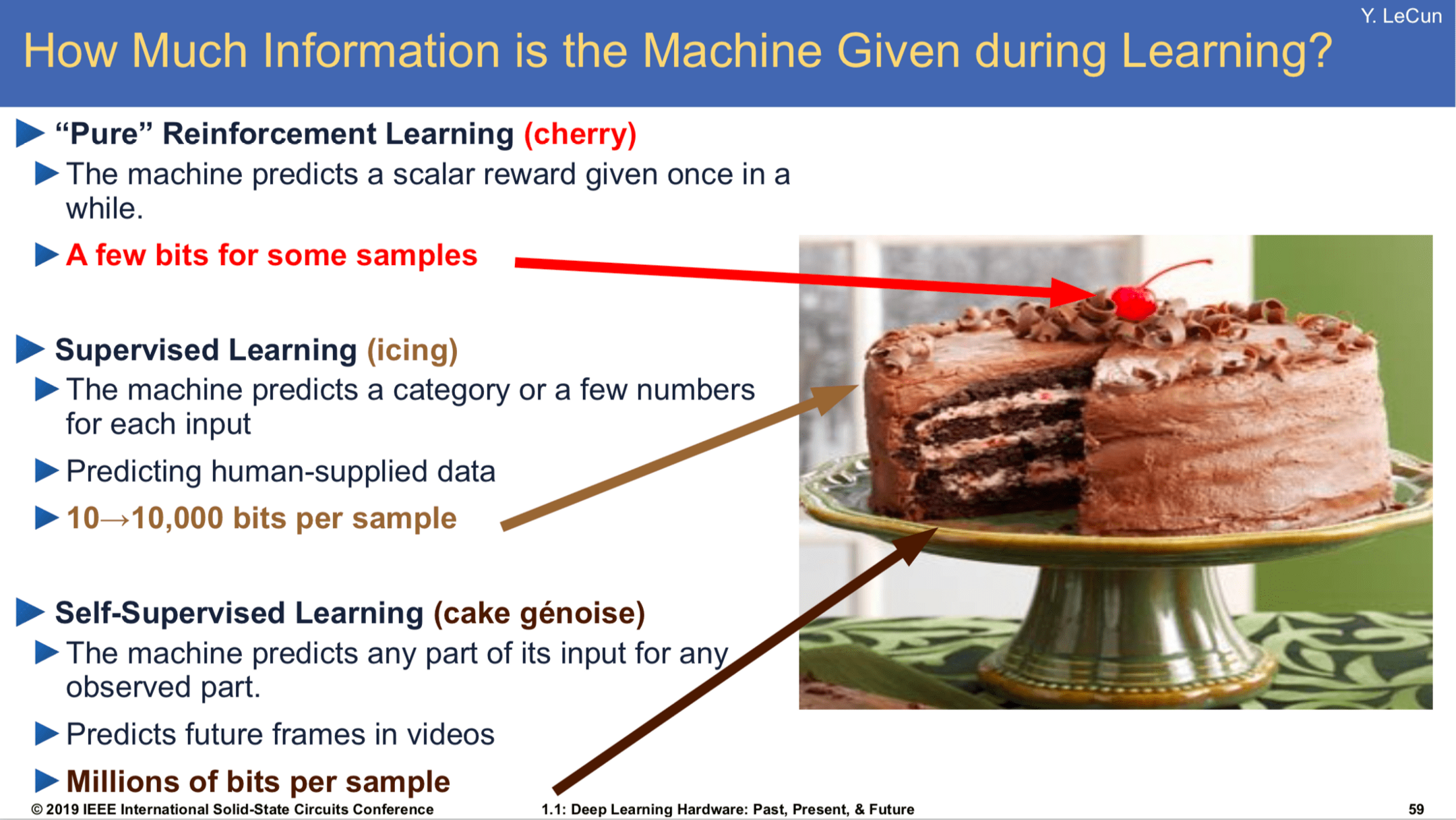}
    \includegraphics[height=0.2448\textwidth]{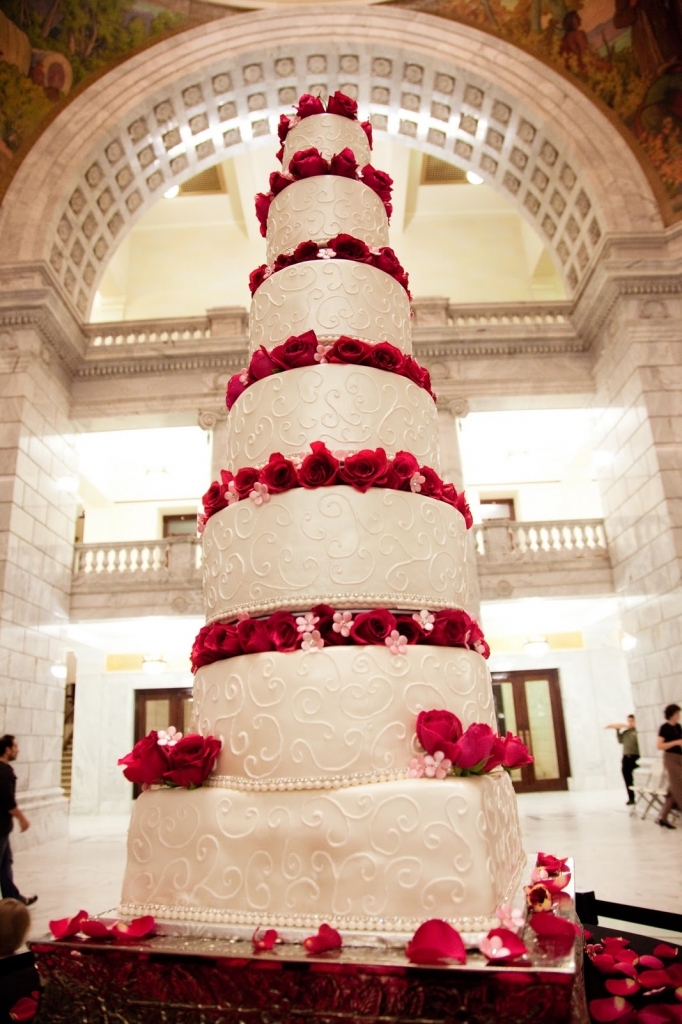}%
    \includegraphics[height=0.2448\textwidth]{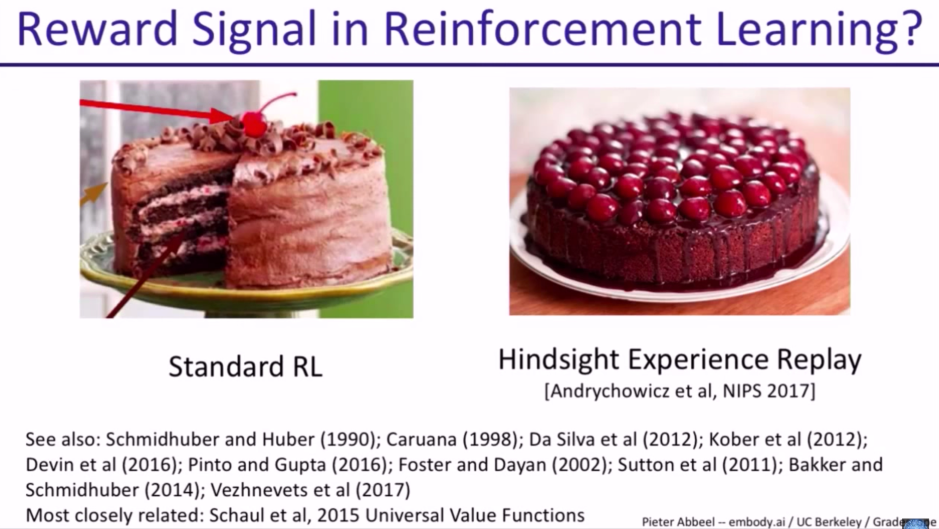}%
    \includegraphics[height=0.2448\textwidth]{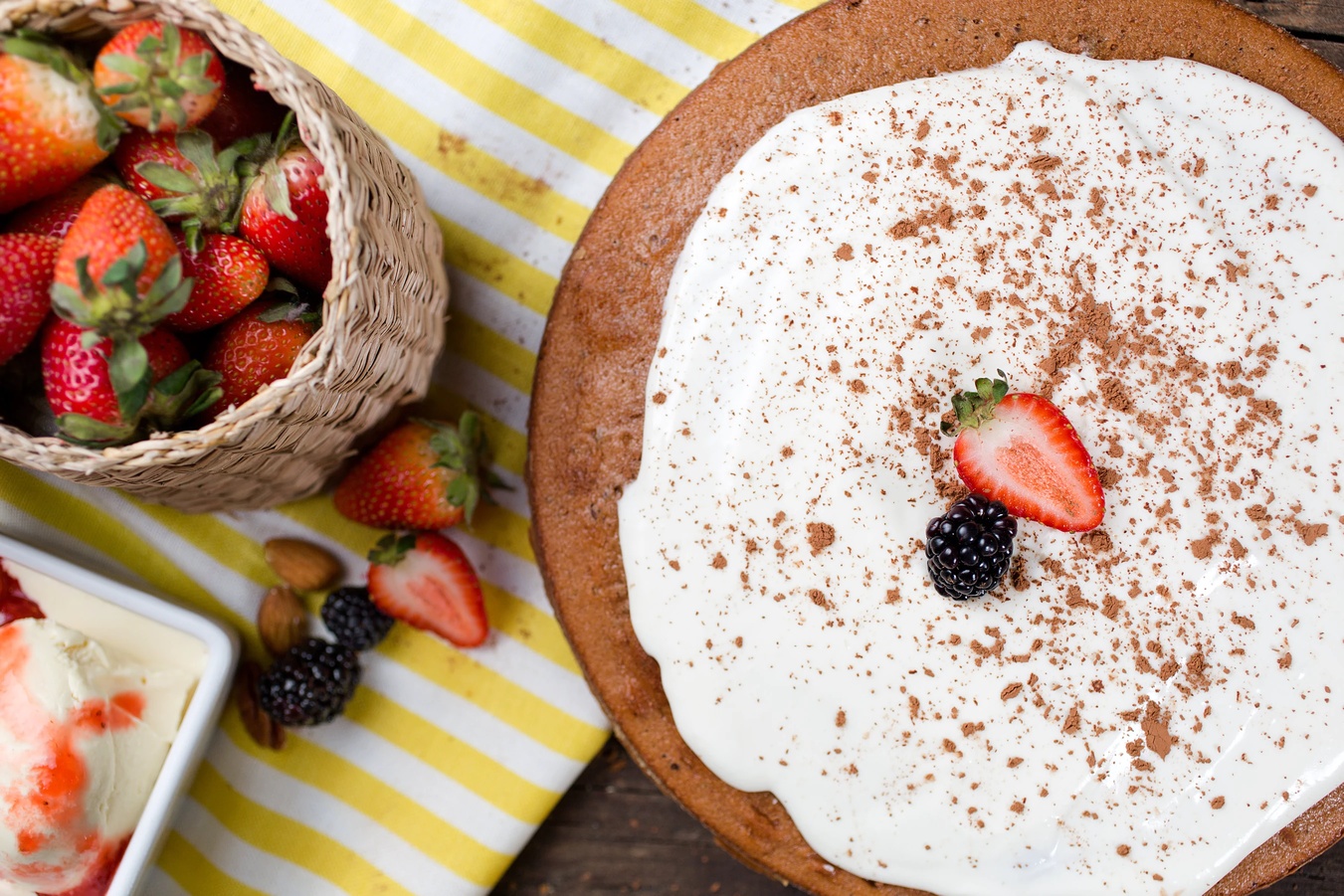}
    \caption[Caption for LOF]{\textbf{Comparison to state-of-the-art unifying theory cake metaphors.} 
    \textit{From top left to bottom right:}
    (i) The seminal multi-layered cake metaphor introduced by~\cite{lecun16}, linking reinforcement learning, supervised learning and predictive learning, (ii) a chef's revision to the base cake, paying homage to the critical role of self-supervised learning~\citep{lecun2019}.
    (iii) a vain attempt to claim state-of-the-tart by simply increasing cake depth~\citep{albanie2018substitute},
    (iv) the hindsight experience replay cake of~\citep{abbeel17}---less unifying than prior work, but with more delicious cherries,
    (v) in this work, we highlight the role of Nature's powerful fourth (hidden) component of learning to complement the three of~\cite{lecun2019}: the ants that evolved to pick up the crumbs of cake that have fallen off the table\footnotemark. A beautiful display of trickle-down eco-nomics. Thus, following the footsteps of famed confectionery enthusiast Marie Antoinette, we shall let them have cake (and by ``them'' we mean all three readers of this article; hello Mrs. Jo\~{a}o).}
    \label{fig:cake}
\end{figure}

\noindent \textbf{Comparison to state of the art.} In Fig.~\ref{fig:cake}, we compare our theory to the existing state of the art in unifying theories, expressed as cake metaphors. Crucially, compared to other unifying theories with at most three components, our theory encompasses not only the cake, but also the ecosystem of nature surrounding it, rendering it comprehensively \textit{more unifying}.

\footnotetext{Image credit: ~\citet{strawberrycake}.}

Having thoroughly validated our framework, we turn next to its implications.  We highlight the critical importance of the \textit{conservation of deep learning models} in ensuring a healthy ML ecosystem for future generations focusing particularly on
experimental conservation efforts.

\noindent \textbf{The Conservation of Deep Learning Models.} Beginning 
with \cite{krizhevsky2012imagenet} there has been a surge of public interest
in neural network architectures. For a time it became
a fashionable practice among high society 
to collect exotic GAN variants, with single-letter-based naming schemes
leading to a quick depletion of both the Latin and
Greek alphabets, and a few failed emoji-based attempts.
In order to satiate this demand, numerous Model Zoos
were established, providing easy access to gigabytes of
model weights and a fun day's activity for the kids.
However, concerns soon arose over the effects of removing
these models from their natural habitats. Models which
were born racing through ImageNet epochs on a 
64 GPU cluster were now being limited to
the cramped and dull confines of an S3 bucket.
Deteriorating conditions at Model Zoos past their
glory days caused further alarm, with ageing \texttt{.caffemodel}
files suffering from protobuf drift and
custom layers lost to time. Eccentric Model Zoo owners
were also known to operate illicit Ganbreeder programmes
supplying the rich and famous.
In the wild, too, many species of models became increasingly
rare and endangered, surviving on only in such
remote corners of research labs as that server
sitting under a PostDoc's desk since 2012 that must
never be unplugged.\footnote{And there it shall remain until the scribbled post-it's glue eventually gives way.}

Organisations such as Big GAN Rescue have sought to
provide sanctuary for old and abandoned models,
operating VMs running MATLAB R2013a and vintage
versions of MatConvNet, allowing these models
to live out the rest of their days with a daily
epoch of vanilla-flavoured CIFAR-10. Efforts have also been directed
towards rewilding, through mass-uploading of models
to peer-to-peer filesharing services, allowing models
to roam across the open plains of the internet as
\texttt{VGG\_16\_BDRIP\_HyPeRDEEP~(fansub).xvid.rar}.
\section{Conclusion} \label{sec:conclusion}

In this article we embarked on an expedition into the far reaches of the digital natural world, and found that much yet remains to be discovered.
There is a vast optimisation landscape, from the tallest Hima-layers, the imposing Mount Foo-ji (used in Python examples worldwide), and the flatlands of the S(a)V(a)N(na), stretching to the bottom of the Mary-Ann trench (so named by Alice and Bob).

\textbf{Acknowledgements.} As a \textit{bona fide} act of self-defensive scholarship, we graciously acknowledge that several sentence fragments were inspired verbatim from the original text of~\cite{darwin1859origin}. To further immunise ourselves from reasonable accusations of plagiarism, we cite big D again here, following a carefully selected number of words after the original citation to maximise efficacy~\citep{darwin1859origin}. We also note that our comprehensive literature review had to be completed before bedtime, and thus should be considered definitive, but not definitively definitive. The authors thank James Thewlis for technical and philosophical support.

\bibliographystyle{iclr_style}
\bibliography{refs.bib}

\end{document}